\documentclass{article}
\usepackage[utf8]{inputenc}

\usepackage{graphicx}
\usepackage{multirow}
\usepackage{subcaption}
\usepackage{array}
\usepackage{yhmath}
\usepackage{url}
\usepackage{epstopdf}
\usepackage{authblk}
\begin{document}

\title{Task-Independent Spiking Central Pattern Generator: A Learning-Based Approach}

\newcommand\Mark[1]{\textsuperscript#1}

\author[1]{Elie Aljalbout}
\author[1]{Florian Walter}
\author[1,2]{Florian R{\"o}hrbein}
\author[1]{Alois Knoll}
\affil[1]{Technische Universit{\"a}t M{\"u}nchen}
\affil[2]{Alfred Kärcher SE Co. \& KG}
\affil[ ]{\textit {\{name.lastname\}@tum.de}}

\date{}


\maketitle

\begin{abstract}
    Legged locomotion is a challenging task in the field of robotics but a rather simple one in nature. This motivates the use of biological methodologies as solutions to this problem. Central pattern generators are neural networks that are thought to be responsible for locomotion in humans and some animal species. As for robotics, many attempts were made to reproduce such systems and use them for a similar goal. One interesting design model is based on spiking neural networks. This model is the main focus of this work, as its contribution is not limited to engineering but also applicable to neuroscience. This paper introduces a new general framework for building central pattern generators that are task-independent, biologically plausible, and rely on learning methods. The abilities and properties of the presented approach are not only evaluated in simulation but also in a robotic experiment. The results are very promising as the used robot was able to perform stable walking at different speeds and to change speed within the same gait cycle.

\end{abstract}

\section{Introduction}
Robotic locomotion is a functionality that enables robots to navigate and transport themselves. One specific form of locomotion is legged locomotion. Despite the challenges faced to achieve this form, it presents multiple advantages such as adaptability to multiple types of terrains (e.g. unpaved ground, rugged terrain, stairs etc.), interaction with the physical environment, agility in terms of generated paths complexity, and superior failure tolerance (e.g. when a leg fails the robot can still move) \cite{machado2006overview}. 

In biology, humans and animals can move and transport their bodies from one place to another in a quite efficient manner, which motivates and highlights the necessity and importance of approaching this problem with more inspiration from biology. In addition, imitating biological locomotion could also lead to more human- and animal-like gaits. However, the functionality leading to biological locomotion is not yet fully understood. Central Pattern Generators (CPGs) are neural networks that are thought to be responsible for multiple rhythmic behaviors, including locomotion. The exact architecture and configuration leading to the ability of these networks to produce rhythmic patterns are still not entirely known. Consequently, this field of research is considered multidisciplinary, since it contributes to both robotics and neuroscience. Another advantage of engineering CPGs is the possibility to implement them on neuromorphic hardware \cite{walter2015neuromorphic}, leading to more power efficiency.

In robotics, many attempts were made to design CPGs for locomotion control. These attempts can be separated into two main groups:
\begin{enumerate}
    \item Methods based on systems of coupled oscillators (SCO)
    \item Spiking neural networks (SNN) based methods
\end{enumerate}
The second type is the one adopted in this work. As for SCOs, they offer the ability to be mathematically analyzed due to their mathematical nature, which made them advance faster than their counterparts that present complex dynamics and harder analytical obstacles. Unfortunately, SCOs lack biological plausibility, which means that their contributions to neuroscience and interventions in neuro-prosthetics are limited. For example, when interfacing prosthetic robotic devices to amputated humans and spinal cord injury patients, SNN-based approaches process and receive the same kind of signals as the CPGs of their biological counterparts \cite{espinal2016design}. The same doesn't apply for SCOs.

In this paper, we introduce a novel SNN-based CGG architecture that combines biological plausibility with easy tunable output behavior. The architecture builds on previously published models but adds three notable features: First, the gait speed can be easily modulated through a tonic input signal even within an ongoing cycle. Second, the generated patterns are not encoded manually or through evolutionary algorithms but directly learned based on a target signal. Third, the number of phases can be defined freely, which, together with the second feature, allows the approach to be general and task-independent. As we will show during the evaluation, this approach is very stable and converges quickly, making it very well-suited for implementing it in robots. Importantly, while also other CPG architectures proposed earlier support speed modulation, pattern learning, or flexible number of phases, none of them supports all of these properties at the same time. Our contribution therefore can both provide valuable insight to both neuroscience-focused CPG research and robot locomotion research.

The rest of this paper is structured as follows. In section~\ref{sec:related}, related methods will be presented, followed by the proposed approach in section~\ref{sec:proposed}. Results and conclusion will be then discussed in sections~\ref{sec:experiments} and \ref{sec:conclusion} respectively.

\section{Related Work}
\label{sec:related}
 Several approaches have been demonstrated to be capable of building CPGs using spiking neural networks. Some of these methods are simply based on hand-tuning of the synaptic efficacies \cite{billard2000biologically,lewis2001control,lewis2005cpg,maufroy2008towards,donati2014spiking,garcia2015central}. Such methods manage to reproduce basic CPG functionalities such as stable walking and in some cases speed control \cite{lewis2005cpg,maufroy2008towards}. The main disadvantage of these approaches is the small size of their networks\footnote{CPGs are known to have a small number of neurons relative to the whole nervous system. The discussed methods use networks that are far smaller than biological CPGs.} which makes the weight tuning process easier but forbids the network from scaling to additional features such as modulation by sensory feedback and autonomous gait transition. Another set of approaches relies on genetic algorithms and similar methods to learn the network weights. For instance, the work in \cite{russell2007configuring} applied genetic evolution on the network from \cite{lewis2005cpg} and succeeded to obtain a similar behavior, omitting the complicated tuning of the locomotion behavior but keeping the same disadvantages as previously discussed. An alternative approach was recently presented in \cite{espinal2016design,espinal2016quadrupedal}, where the CPG network is formulated as a Christiansen grammar and an evolutionary process is performed on it. Interestingly, this method not only learns the network's weights but also its architecture, which is a great advantage offering more automation for the CPG building process. Further approaches based on evolutionary algorithms were also proposed in \cite{orchard2008configuring} and \cite{caamano2008using}. Besides the obvious disadvantage of slow convergence, evolutionary methods don't scale to larger networks and their training method is not biologically plausible, which also limits their neuroscientific contribution. Other methods also fall in the category of non-plausible to biology such as \cite{smolinski2015automated}, where it is proposed to perform brute-force search on the CPG parameters space, and \cite{kuroe2006learning} where a gradient descent based method is presented. These methods have multiple disadvantages such as slow convergence (especially in the case of brute-force) and being incapable to scale to advanced CPG features. Finally, the method in \cite{ponulak2006adaptive} presents a general framework to build task-independent CPGs. It relies on the remote supervision method (ReSuMe) \cite{ponulak2005resume} to learn the weights connecting two populations of randomly connected neurons to the motor neurons of the CPG. This method uses biologically plausible learning but a non-plausible network architecture. Additionally, it only achieves stable walking at a fixed speed and omits additional CPG features.

\begin{figure}
    \centering
    \includegraphics[width=\textwidth]{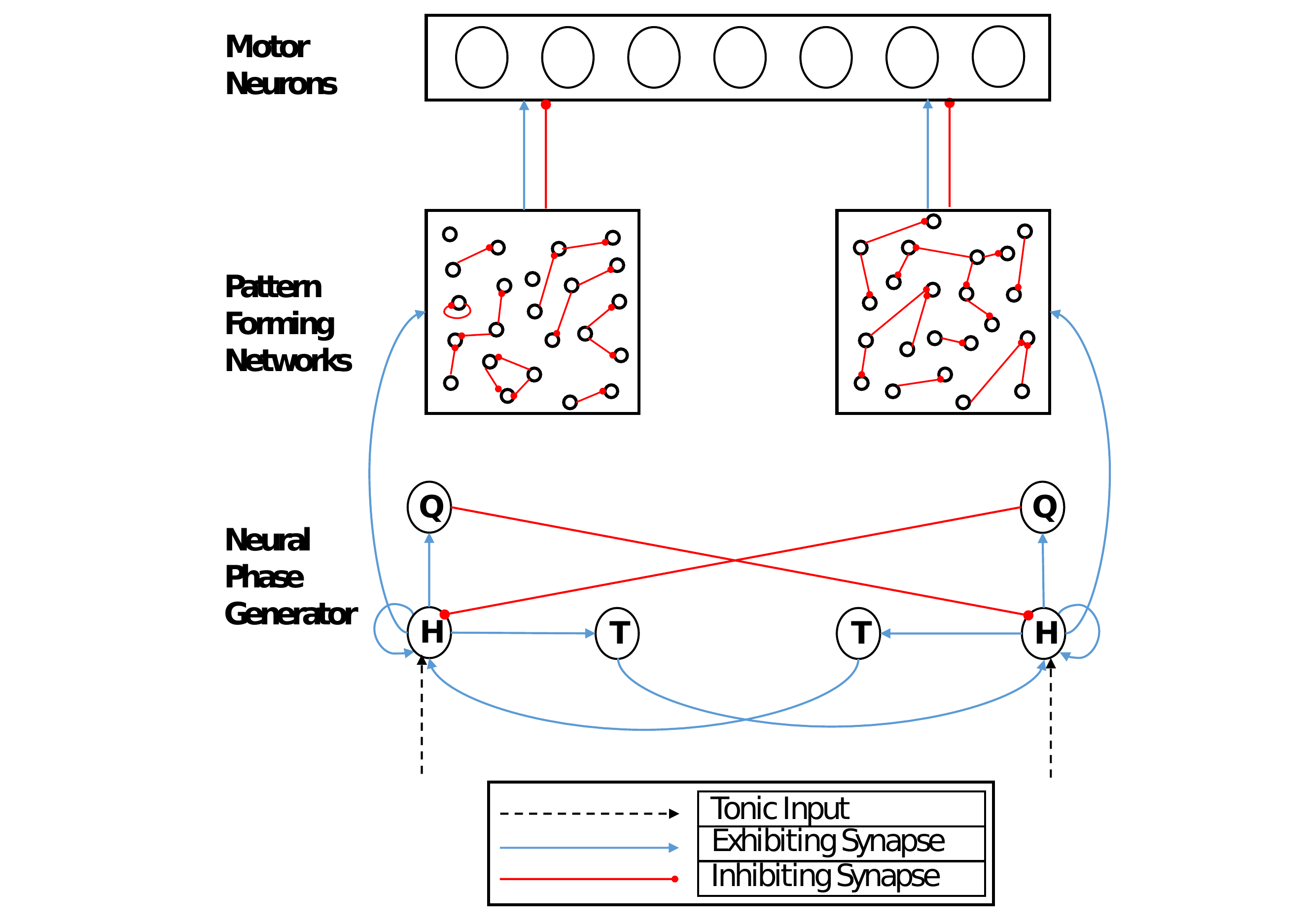}
    \caption{The proposed network architecture for a CPG with two phases. The neural phase generator (NPG) generates a rhythm corresponding to the number of desired phases. The H neuron is the representative of the activity of each phase, Q takes care of inhibiting other phase modules and T is responsible for the transition to another phase. This rhythm is then propagated to the pattern forming networks (PFN) which map the NPG outputs to a higher dimensional space. PFNs output is then fed to motor neurons which activate the robot's muscles.}
    \label{fig:generalarch}
\end{figure}

\section{Proposed Approach}
\label{sec:proposed}

As previous methods either relied on hand-tuning, neglected advanced CPG features, or couldn't scale to larger CPG networks, we present the following approach, which is the first to successfully combine all of these characteristics using a biologically plausible network architecture. From a general perspective, the architecture consists of three main components. The first one is the neural phase generator (NPG), which is adapted from \cite{maufroy2008towards} with minor modifications. In analogy to the biological model in \cite{rybak2006modelling}, this part plays the role of the rhythm generator. The second part is the pattern forming network (PFN), which consists of networks with random parameters. At this level spiking patterns appear in a rhythmic fashion with respect to the NPG's output. Finally, the last part is the pool of motor neurons, which activates the actuated muscles through spike activity. Each of these components will be discussed next in more details.

Figure \ref{fig:generalarch} shows the proposed architecture in the case of a CPG which encodes only two phases. For this reason, the architecture is almost symmetric, when omitting the randomness in the PFNs. In the case when more phases are required, the architecture is extended in all three parts. For instance, at the NPG level, more modules should be added corresponding to the new requirement (Figure \ref{fig:npgarch}). Similarly, at the second level, random networks are also created with respect to the number of desired phases. Finally, in the pool of motor neurons, the number of neurons can also be different but not necessarily. In the following, each of these components will be described separately, in terms of their inner connections, functionality, and role.

\subsection{Neural Phase Generator}

\begin{figure}
    \centering
    \includegraphics[width=\linewidth]{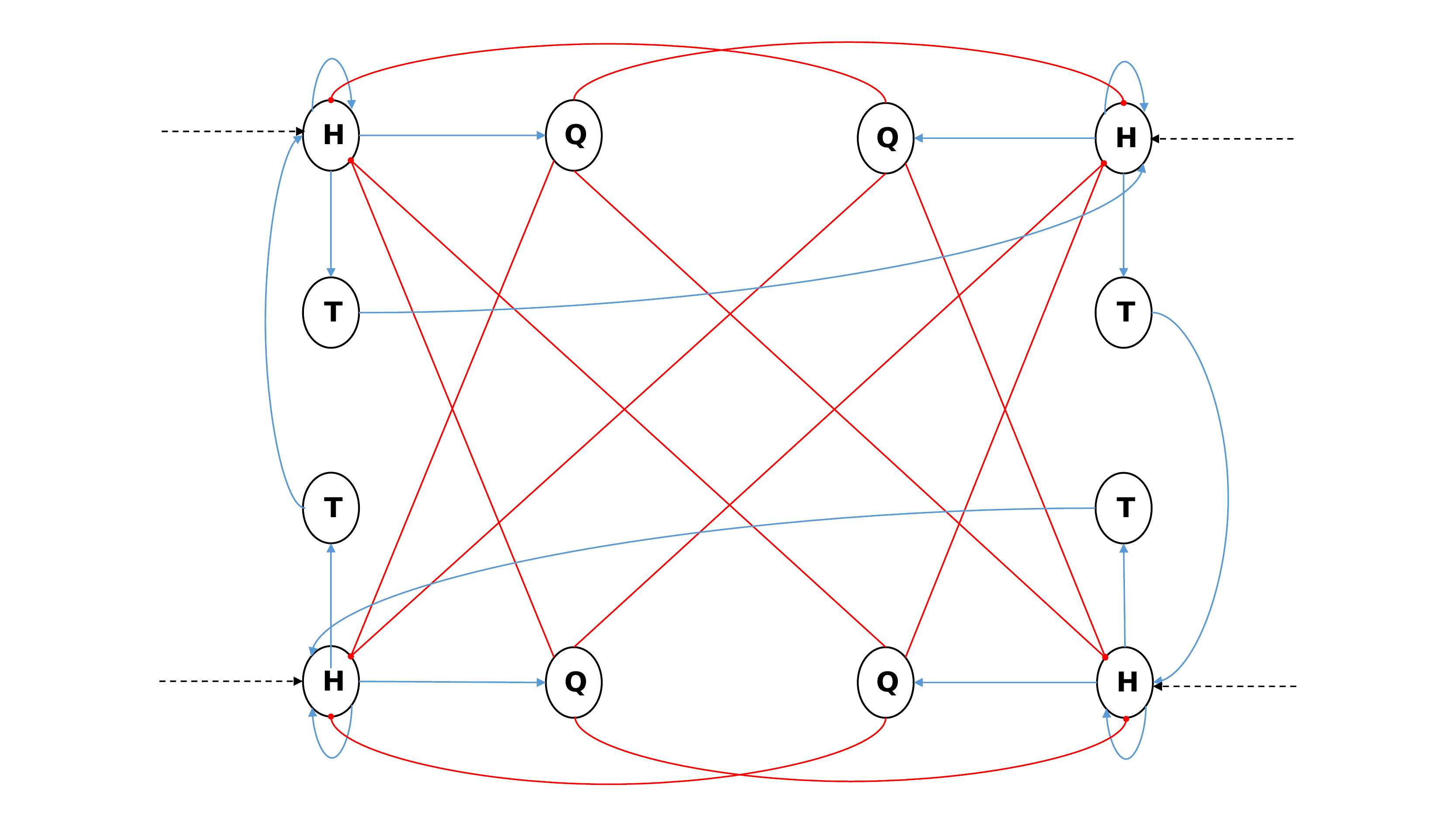}
    \caption{Illustration of an NPG network for the case of a CPG having four phases.}
    \label{fig:npgarch}
\end{figure}

As already mentioned, this part is adapted from \cite{maufroy2008towards}. Briefly, each NPG module consists of three neurons H, Q and T (as seen in Figure \ref{fig:npgarch}). The H neuron is the representative of the module's activity. In other words, whenever a module's H neuron is spiking, the phase this module represents is active. As for the Q neuron, its role is to guarantee that other NPG modules are inactive when required. It does so by inhibiting the H neurons of the corresponding modules. As for T neurons, they are responsible for the transition from module to module. In contrast to the NPG in \cite{maufroy2008towards}, our modified version receives an external tonic input, which is responsible for the start and end of activity by exciting and inhibiting the neurons of the NPG. This input represents the modulating inputs issued from the brain to control the CPG's activity. Once the NPG is active, cutting the external input won't stop its activity. Instead, the network will keep oscillating because of its inner mutual connections. This behavior is compatible with the nature of CPGs being able to oscillate even in the absence of input from the brain. To make sure this property is available, a constraint is required: the auto-synapse of the H neurons should have weights that are high enough to ensure a continuous spiking behavior until the respective T neurons fire, which then results in a phase transition. Moreover, when the tonic input has a higher frequency, the H neuron will fire more frequently resulting in a faster transition to the next phase. This modification gives control to the NPG over the speed and phase properties of the produced gaits. This behavior is plausible with the proposed biological model where this control is available at the level of the rhythm generator. In \cite{maufroy2008towards}, this ability is given to the second layer of the architecture which in their case is the motor output shaping stage. Compared to our architecture this would correspond to giving control to the PFNs.

Another modification, to the NPG of \cite{maufroy2008towards}, is in the number of T neurons in each module. In the previous work, two T neurons were present in every module, as for ours, this number is variable. It is dependent on the duration of the phase each module is responsible for, the longer the duration the bigger the number. All T neurons except the last one, receive the same external tonic input (as the H neuron), which is important since they each excite a part of the PFN. However, these intermediate T neurons, in addition to their tonic inputs are not visible in figure \ref{fig:npgarch} for the sake of simplicity.

Finally, the number of modules in our version of the NPG is not fixed to two, but instead is flexible depending on the number of desired phases, as illustrated in figure \ref{fig:npgarch}. This flexibility allows the architecture to be general and adaptable for different types of locomotion.

\subsection{Pattern Forming Networks (PFN)}
\label{section:pfn}
This part of the architecture consists of several separate networks. The number is equivalent to the number of phases of the desired CPG. Each of these networks corresponds to a module of the NPG and is excited by its H and intermediate T neurons. The neurons within each of these networks have randomly initialized properties, leading to richer dynamics and higher learning abilities.

In order to simplify the learning procedure, a minor restriction is enforced: neurons of the PFNs are required to spike only once during each CPG cycle. This is achieved by creating another network with a similar number of neurons for every PFN. This second network is named inhibiting network (IN). 
\begin{figure}
\centering
\includegraphics[width=0.8\linewidth]{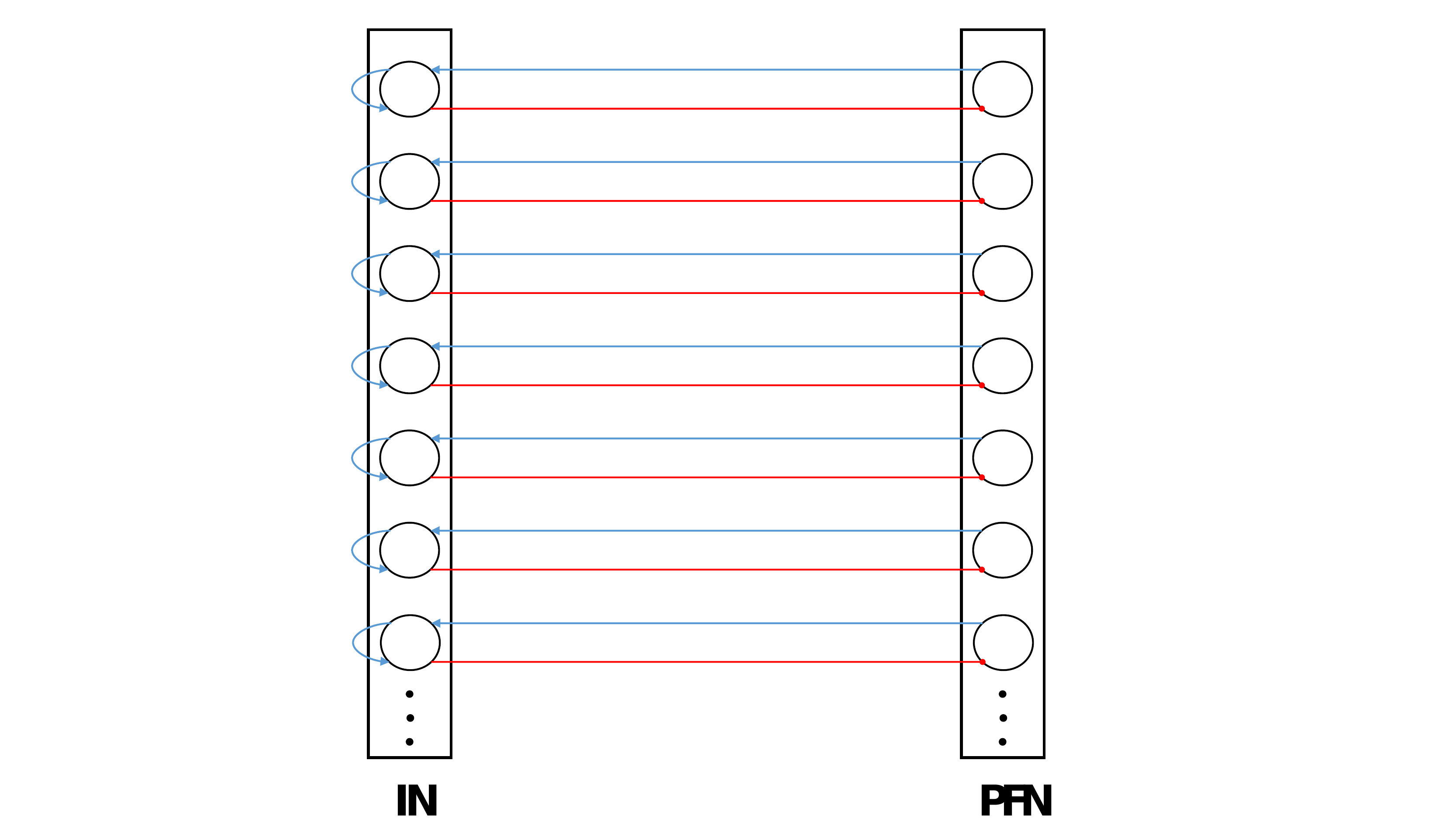}
\caption{Illustration of inhibiting networks. Neurons in the left column are the inhibiting neurons, and the ones on the right are those of the PFN.}\label{fig:inhnet}
\end{figure}
Both networks are then connected in a one-to-one fashion with excitation issued from the PFN to the IN and with inhibition in the opposite direction. INs have no synapses connecting their neurons, but instead, exciting auto-synapses for all of their neurons to ensure that each one of them keeps spiking once it's activated. To illustrate how this works: each time a neuron in a PFN spikes, it'll excite a corresponding neuron in an IN leading it to spike in a continuous fashion and inhibit reciprocally the same PFN neuron that made it fire. This concept is illustrated in figure \ref{fig:inhnet}. Now to make sure that the IN is reset in the next cycle, it'll be inhibited by a Q neuron of an NPG module from a different phase.

Additionally, neurons within a certain PFN are connected with random mutual inhibition at a $10\%$ rate of connectivity (i.e. the number of synapses is $10\%$ the one of an all-to-all connectivity scheme), which ensures that spikes produced by different neurons within the network are well distributed over the whole duration of the corresponding phase, and not concentrated in a small duration right after the NPG spikes. Distribution of the PFN spikes in time is very important for the learning, in a way that the more spikes are distributed, the more the number of desired behaviors that can be learned increases. The intuition behind that is simple. To illustrate it, consider the case where no spike occurs in the time interval $[t,t+\epsilon]$ at the PFN. This will make it less probable that any spike occurs in this same interval in the pool of motor neurons, which are only excited by PFN neurons. In conclusion, PFN spikes should be the most possibly scattered over the phase duration. To enforce that, the size of the network can be increased in order to obtain a higher spiking rate. Conceptually, the role of the PFN can be seen as mapping the NPG spikes into a higher dimensional space in which spikes are well distributed in time. Finally, the external control inputs of the NPG are also propagated to this layer. When the frequency of these signals is increased, the frequency of spiking in the H and intermediate T neurons in the NPG will also increase. Thus, the neurons of the PFN will fire sooner, which means that also the speed of gait will increase. The relation between the frequency of spiking at the NPG level and the time of spikes in the PFN is not a perfect mapping, but that is left for the learning to figure out. Concerning the learning, it is important to note that it's dependent on the PFN's size. For instance, the larger the PFN is, the more probable it is that the learning method will converge. However, the convergence will then require longer training time. Throughout our experiments, we used PFNs with sizes ranging from $150$ to $500$ neurons. The exact selection of the number is also dependent on each phase's duration. 
\subsection{Pool of Motor Neurons}
\label{poolmn}
\begin{figure}
    \centering
    \includegraphics[width=\linewidth]{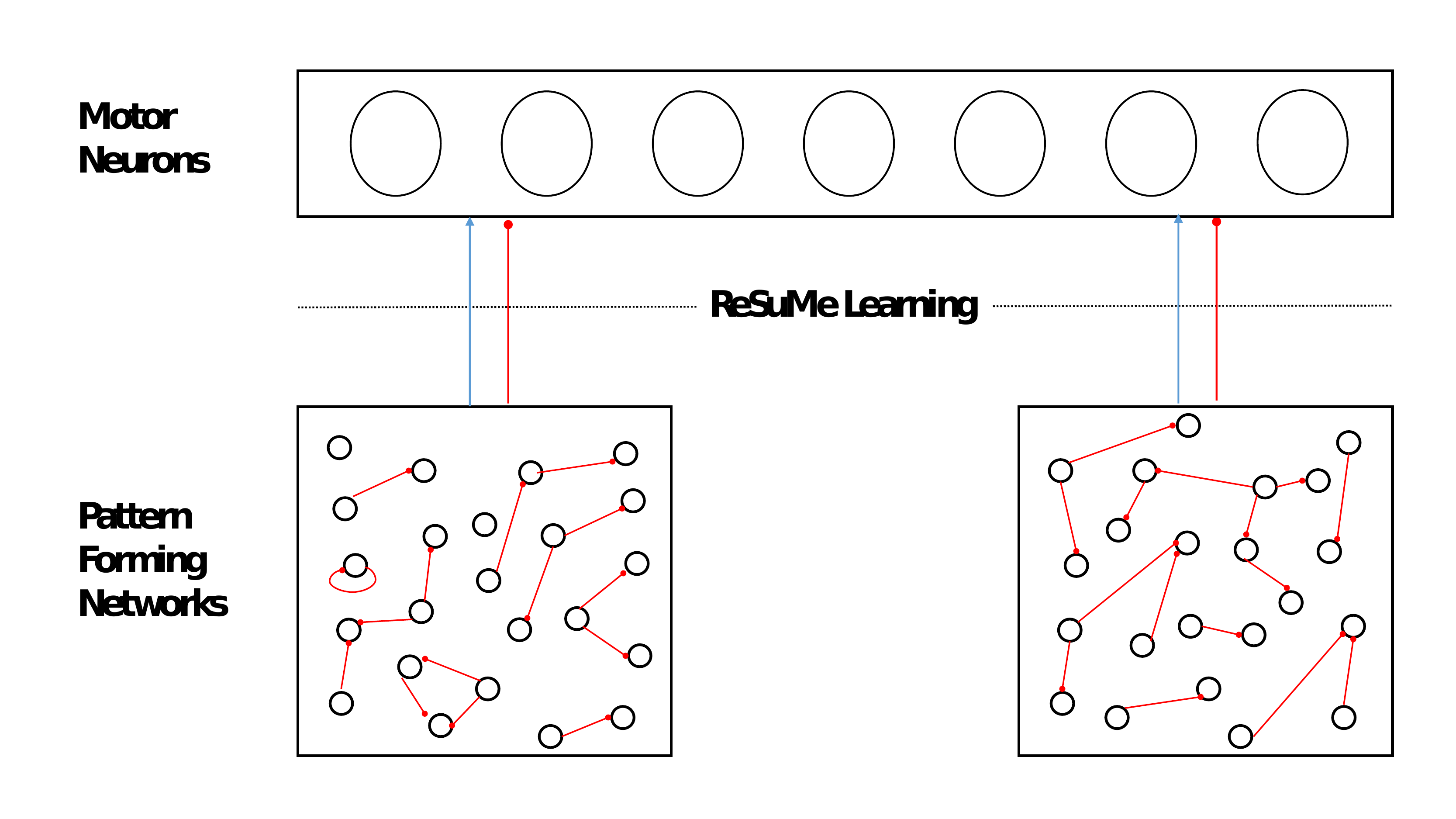}
    \caption{Pool of motor neurons, which receive excitatory and inhibitory inputs from the PFN. The weights of the synapses connecting PFNs and these neurons are learned using the remote supervision method ReSuMe.}
    \label{fig:mnarch}
\end{figure}

This last part of the architecture is where rhythmic patterns really matter, since the spikes at this level are responsible for muscle activation and producing gaits for legged locomotion. In biology, the precise way these spikes are decoded into muscle activation is not fully known, but research has indicated that population rate coding is used \cite{gerstner1997neural,maunsell1983functional}. In other terms, the activation of a muscle is encoded by a population of neurons. Their spike rate is related to the level of activation of the muscle. Therefore, the number of neurons in the pool is dependent on the desired locomotion and the number of actuated muscles. A similar scheme was adopted in the robotic experiment performed within this work. Moreover, neurons within the pool are not inter-connected, but receive as input the spikes produced by the PFNs, through synapses of both excitatory and inhibitory types. The weights of those synapses are the only ones that are learned. They are adapted to obtain the desired locomotion behavior. The learning procedure will be discussed later. Finally, the control signals (external tonic input) applied at the NPG are also propagated to the neurons in this pool. For instance, when the tonic input has an increased frequency, the speed of the spike pattern produced by the PFN will increase. Ideally, this will lead to the spiking behavior of the motor neurons to be accelerated.

\subsection{Learning}
 When designed faithfully to the constraints already mentioned, the previously described architecture is capable by itself of producing rhythmic spiking patterns in the pool of motor neurons. However, this rhythmic spiking behavior is random and does not correspond to any locomotive behavior. Therefore, the network should be trained somehow to adapt to the desired spiking behavior\footnote{The desired spiking behavior is specified by the CPG designer and can also be given by demonstration.} in order for it to be used in real robot locomotion tasks. In order to do so, the Remote Supervision Method (ReSuMe) \cite{ponulak2005resume}, was employed. However, the only synapses of the network which learning is applied on, are the ones connecting the PFNs to the pool of motor neurons. 
 
 The way ReSuMe works is as follows \cite{ponulak2005resume}: For each of the learning neurons (in our case the motor neurons), a teacher signal is associated with predetermined timing, representing the desired spiking behavior. This signal is not delivered to the learning neuron, but still plays an important role in the learning updates of synapses terminating at it. Moreover, the modification is based on two rules, the first one depends on the correlation between the presynaptic and the desired spike times, as for the second, it depends on the correlation between the presynaptic and the postsynaptic spike times~\cite{ponulak2005resume}. The following is the corresponding modification function for the synaptic weight between a presynaptic neuron $k$ (from the PFN) and a postsynaptic neuron $i$ (from the motor neurons):
 
 \begin{equation}
\label{eq:resume1}
\begin{split}
    \frac{d}{dt}w_{ki}(t)=&S^d(t)\Big[a^d+\int_{0}^{\infty}W^d(s^d)S^{in}(t-s^d)ds^d  \Big]\\
    +&S^l(t)\Big[a^l+\int_{0}^{\infty}W^l(s^l)S^{in}(t-s^l)ds^l  \Big]
\end{split}
\end{equation}

Where $S^d$, $S^l$ and $S^{in}$ are respectively the target, the post and the presynaptic spike trains, $a^d$ and $a^l$ determine the so-called non-Hebbian processes of weight modifications, and $s^l$ and $s^d$ represent respectively the difference between the time of spike of the postsynaptic neuron and the presynaptic one and the difference between the time of the teacher signal and the time of spike of the presynaptic neuron. In the case of excitatory synapses, the terms $a^d$ and $a^l$ are positive, and negative otherwise. As for $W^d$ and $W^l$, they represent the learning windows, and are formulated as follows:
\begin{equation}
\label{eq:resumewd}
    W^d(s^d)=\begin{cases}
    \ +A^d\cdot exp(\frac{-s^d}{\tau{}^d}),& \text{if } s^d>\, 0 \\
    0,              & \text{if } s^d\leq0
\end{cases}
\end{equation}
\begin{equation}
\label{eq:resumewl}
    W^l(s^l)=\begin{cases}
    \ -A^l\cdot exp(\frac{-s^l}{\tau{}^l}),& \text{if } s^l>\, 0 \\
    0,              & \text{if } s^l\leq0
\end{cases}
\end{equation}
Where $W^d$ and $W^l$ correspond respectively to the learning window of the target and postsynaptic neurons. As for $A^d$, $A^l$, $\tau^d$ and $\tau^l$, they are all constants, such that $A^d$, $A^l$ are positive for excitatory synapses and negative otherwise and $\tau^d$, $\tau^l$ are always positive.
When setting $a^d=-a^l=a$, $\tau^d=\tau^l$ and $A^d=A^l$, then equation~(\ref{eq:resume1}) takes the following form:
\begin{equation}
\label{eq:resume}
    \frac{d}{dt}w_{ki}(t)=[S^d(t)-S^l(t)]\Big[a^d+\int_{0}^{\infty}W^d(s)S^{in}(t-s^d)ds^d  \Big]
\end{equation}Intuitively, what the method does to the network is to strengthen (weaken) synapses which are excitatory (inhibitory) and are incoming to a certain motor neuron, when a spike is transmitted through these synapses within the learning window as defined by $\tau$, whenever the subject motor neuron is desired to spike. This rule pushes the motor neurons to spike at the desired times. Additionally, it weakens (strengthens) synapses which are excitatory (inhibitory) and are incoming to a certain motor neuron when a spike is transmitted through these synapses within the learning window defined by $\tau$, whenever the motor neuron actually spikes. As for this rule, it forbids motor neurons from spiking at undesired times.
 
\section{Experiments}
\label{sec:experiments}
We designed experiments to answer the following questions:
\begin{itemize}
    \item To which extent is the presented approach capable of learning desired spiking patterns?
    \item Is the targeted speed modulation property working after the learning procedure is done?
    \item Can the system control the locomotion of legged robots?
\end{itemize}
For the first two questions, we designed experiments in a simulation-only environment as described in section~\ref{sec:simulated}. Additionally, we built a real-world experiment with a legged robot (section~\ref{sec:robotics}) to answer the last question and to validate the second one again. The setup for both experiments is described in section~\ref{sec:setup}.

\begin{table}[]
\resizebox{\textwidth}{!}{
\begin{tabular}{l|clclclclclcl}
\hline
\multicolumn{1}{c|}{Properties} & \multicolumn{2}{c}{\begin{tabular}[c]{@{}c@{}}External\\  input \\ current\\  (pA)\end{tabular}} & \multicolumn{2}{c}{\begin{tabular}[c]{@{}c@{}}Resting \\ membrane \\ potential \\ (mV)\end{tabular}} & \multicolumn{2}{c}{\begin{tabular}[c]{@{}c@{}}Capacity\\  of the \\ membrane \\  (pF)\end{tabular}} & \multicolumn{2}{c}{\begin{tabular}[c]{@{}c@{}}Membrane \\ time \\ constant \\ (ms)\end{tabular}} & \multicolumn{2}{c}{\begin{tabular}[c]{@{}c@{}}Spike \\ threshold \\ (mV)\end{tabular}} & \multicolumn{2}{c}{\begin{tabular}[c]{@{}c@{}}Reset \\ potential \\ (mV)\end{tabular}} \\ \cline{2-13} 
 & \multicolumn{1}{l}{min} & max & \multicolumn{1}{l}{min} & max & \multicolumn{1}{l}{min} & max & \multicolumn{1}{l}{min} & max & \multicolumn{1}{l}{min} & max & \multicolumn{1}{l}{min} & max \\ \hline
CPG & \multicolumn{2}{c}{0.0} & \multicolumn{2}{c}{-70.0} & \multicolumn{2}{c}{250.0} & \multicolumn{2}{c}{10.0} & \multicolumn{2}{c}{-55.0} & \multicolumn{2}{c}{-70.0} \\
PFN & \multicolumn{1}{l}{0.0} & 150.0 & \multicolumn{1}{l}{-90.0} & -70.0 & \multicolumn{1}{l}{100.0} & 300.0 & \multicolumn{1}{l}{9.0} & 30.0 & \multicolumn{1}{l}{-50.0} & -30.0 & \multicolumn{1}{l}{-90.0} & -60.0 \\
Motor & \multicolumn{2}{c}{0.0} & \multicolumn{2}{c}{-70.0} & \multicolumn{2}{c}{250.0} & \multicolumn{2}{c}{10.0} & \multicolumn{2}{c}{-55.0} & \multicolumn{2}{c}{-70.0} \\ \hline
\end{tabular}
}
\caption{Parameters of the spiking neurons used in the experiments.}
\label{tab:params}
\end{table}

\subsection{Setup}
\label{sec:setup}

In order to allow our results to be reproducible, we provide the implementation details we used along with our experiments. Namely, we used the NEST simulator \cite{Gewaltig:NEST}. In all of our experiments, the leaky integrate-and-fire neuron model \cite{abbott1999lapicque} was used with alpha-function shaped synaptic currents. As shown in Table~\ref{tab:params}, all neurons used the NEST default parameters except the ones of the PFN, which are initialized with random uniform values. Parameters that are not mentioned in Table~\ref{tab:params} are set to default. As for synapses, they had different initialization ranges at different levels. At the PFN level, neurons were connected to each other with synapses initialized with uniformly random weights between $-3$ and $-1$. Neurons of the PFN and motor neurons are connected together with an all-to-all connectivity pattern. We experimented with several numbers of PFN neurons between $300$ and $3000$ and all values worked out. It is important to note that a larger number of neurons allows a smaller spike shift error, however leads to a longer training time. $20\%$ of the total number of neurons at this level are inhibitory and have initial weights initialized with uniform random values between $-25$ and $-1$. The rest of the neurons are excitatory and have their weights initialized with uniform random values between $1$ and $5$. 

As for the learning, we tried several combinations of hyper-parameters for ReSuMe and report the following ones as being the best in terms of speed and accuracy of learning. For learning windows $W^d$ and $W^l$ we used $3ms$. Additionally we found out that $a=3$, $A^d=A^l=6$ and $\tau_d=\tau_l=2$ perform best.

\subsection{Simulated experiments}
\label{sec:simulated}

\newcommand*{\addheight}[2][.5ex]{%
  \raisebox{0pt}[\dimexpr\height+(#1)\relax]{#2}%
}

\begin{figure}
\centering
\begin{tabular}{cc}
\addheight{\includegraphics[width=0.47\linewidth]{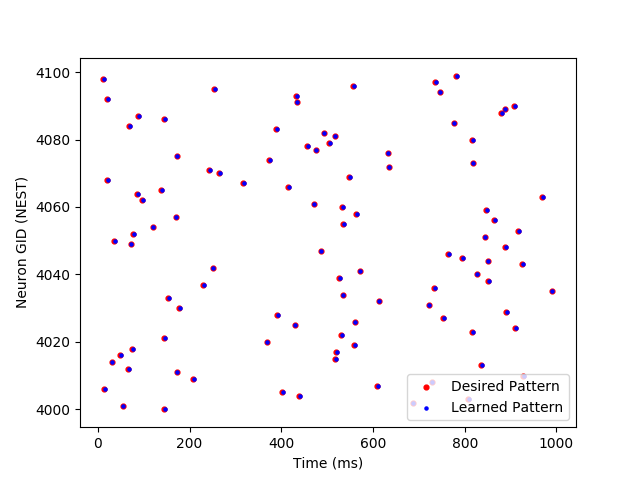}} &
      \addheight{\includegraphics[width=0.47\linewidth]{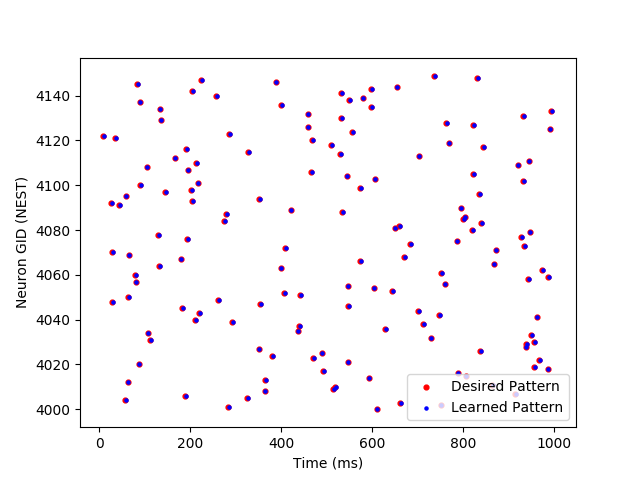}} \\
      \small $s=100$ & $s=150$ \\
      \addheight{\includegraphics[width=0.47\linewidth]{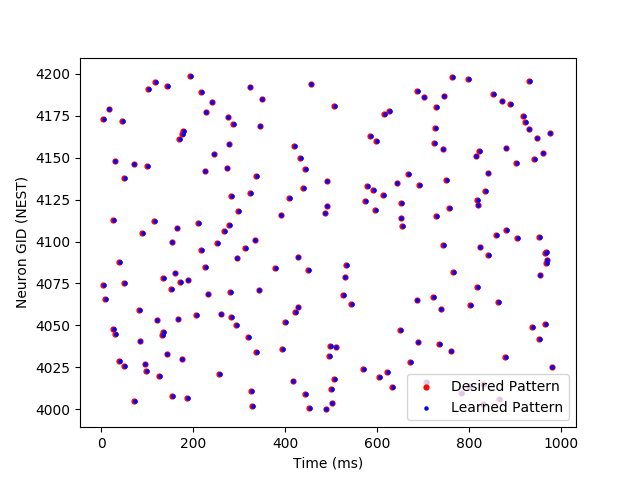}} &
      \addheight{\includegraphics[width=0.47\linewidth]{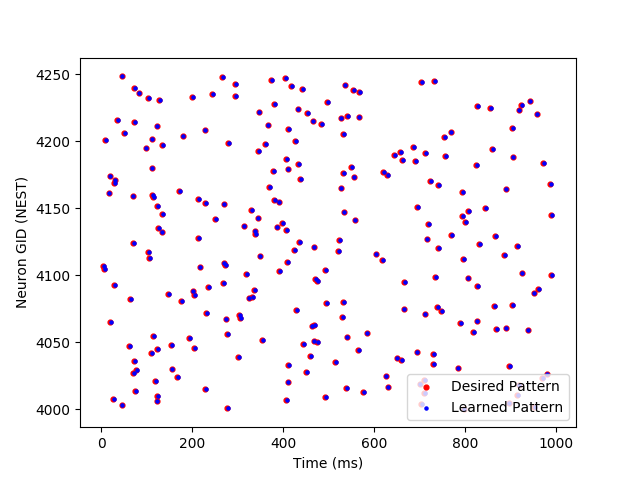}} \\
      \small $s=200$ & $s=250$ 
\end{tabular}
\caption{Spiking patterns obtained for CPGs with $s$ motor neurons for a single CPG cycle. Red dots represent desired spiking patterns randomly generated during the testing phase. Blue dots represent the patterns learned using the presented method.}
\label{fig:adaptiveresults}
\end{figure}

In order to initially validate the proposed method, we tested it in simulation. Namely, we applied the method to more than $250$ randomly generated target spiking sequences with different numbers of motor neurons and let the network learn to imitate these sequences. We experimented with pools of motor neurons of sizes ranging from $5$ to $250$. As a result, the method managed to successfully learn how to imitate all presented spiking sequences without any need for parameter hand-tuning. The average spike shift error of all experiments was very low ($\tt{\sim} 2.1ms$). Figure~\ref{fig:adaptiveresults} shows four different spiking patterns that have been successfully learned at the level of the motor neurons. These patterns are generated by our CPG within a single cycle. These results show that the proposed architecture together with ReSuMe enables learning any desired locomotive behavior with minimal design effort. Consequently, it is possible to embed any desired locomotive gait (whether measured from an already existing walking system or self-defined) in a spiking neural network.

\begin{figure}
\centering
\begin{tabular}{cc}
\addheight{\includegraphics[width=0.47\linewidth]{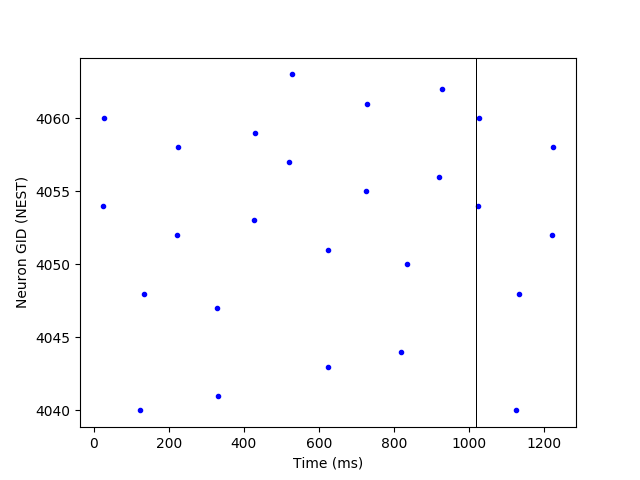}} &
      \addheight{\includegraphics[width=0.47\linewidth]{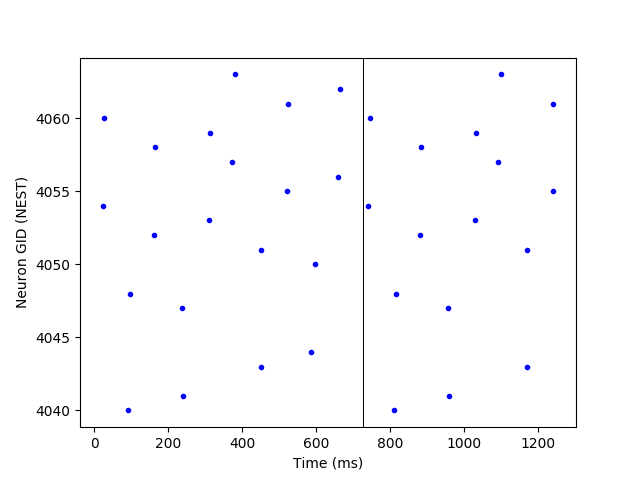}} \\
      \small $s=25$, $F=250$ & $s=25$, $F=500$ \\
      \addheight{\includegraphics[width=0.47\linewidth]{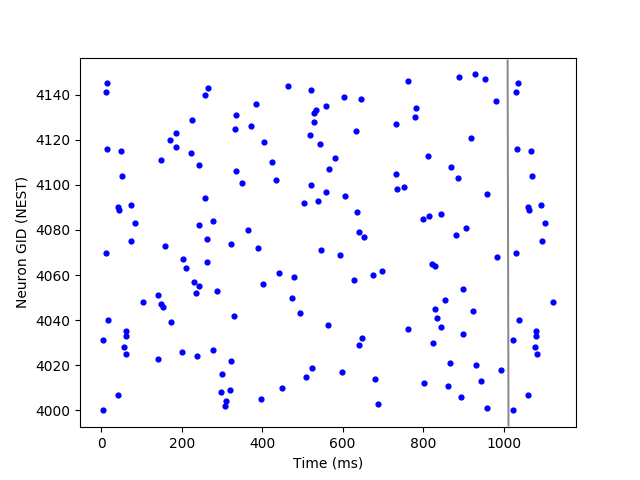}} &
      \addheight{\includegraphics[width=0.47\linewidth]{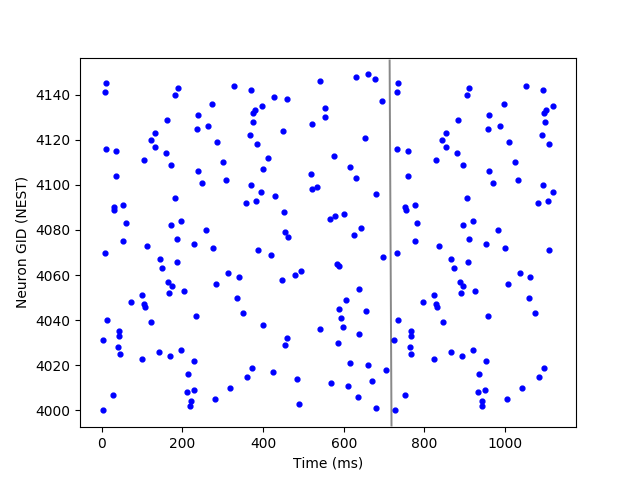}} \\
      \small $s=150$, $F=250$ & $s=150$, $F=500$
\end{tabular}
\caption{Speed modulation of two CPGs with different number of motor neurons $s$ in  each row. When increasing the spiking frequency $F$ of the tonic input from 250 spikes/s to 500 spikes/s we obtain an increased speed in the motor output. Vertical lines indicate the end of a spiking cycle.}
\label{fig:proposedresultsl}
\end{figure}

Additionally, the obtained CPGs are capable of acceleration and deceleration of the same spiking sequences when presented with a higher and lower frequency tonic inputs respectively. Figure~\ref{fig:proposedresultsl} illustrates this property. In the first row, a CPG with a small number of motor neurons is shown, as it's easier to watch the modulation property with a sparse plot. However, our method is also capable of modulating larger numbers of motor neurons as shown in the second row. In most cases, such results were obtained by training the CPG network using only one tonic input frequency. However, in some rare cases, speed modulation was only obtained after training the network under different tonic input frequencies.

\begin{figure}
    \centering
    \begin{subfigure}{0.5\textwidth}
        \centering
        \includegraphics[width=\linewidth]{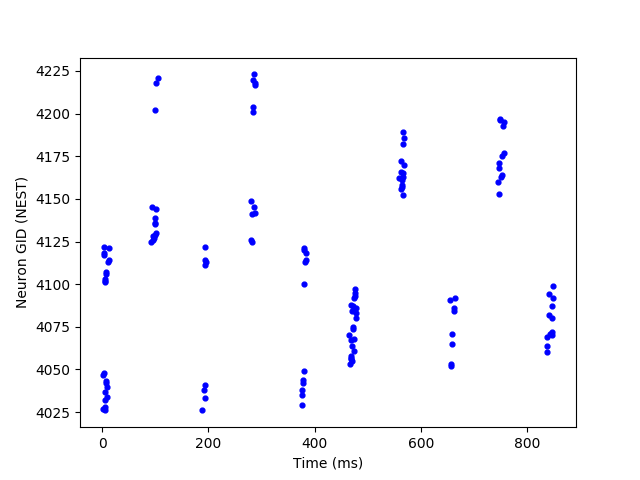}
        \caption{$F=250$ $spikes/s$}
    \end{subfigure}%
    \begin{subfigure}{0.5\textwidth}
        \centering
        \includegraphics[width=\linewidth]{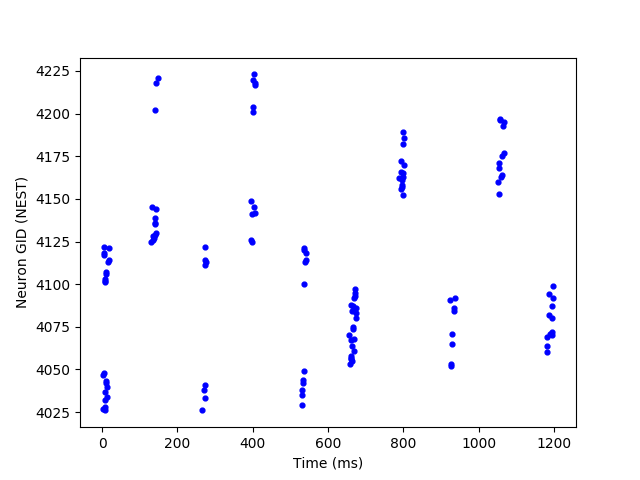}
        \caption{$F=500$ $spikes/s$}
    \end{subfigure}
    \caption{Outputs of the motor neurons used in the robot experiment under different tonic input frequencies $F$.}
    \label{fig:robotexp}
\end{figure}
\subsection{Robotics experiment}
\label{sec:robotics}
\begin{figure}
    \centering
    \begin{subfigure}{0.45\textwidth}
        \centering
        \includegraphics[width=0.97\textwidth]{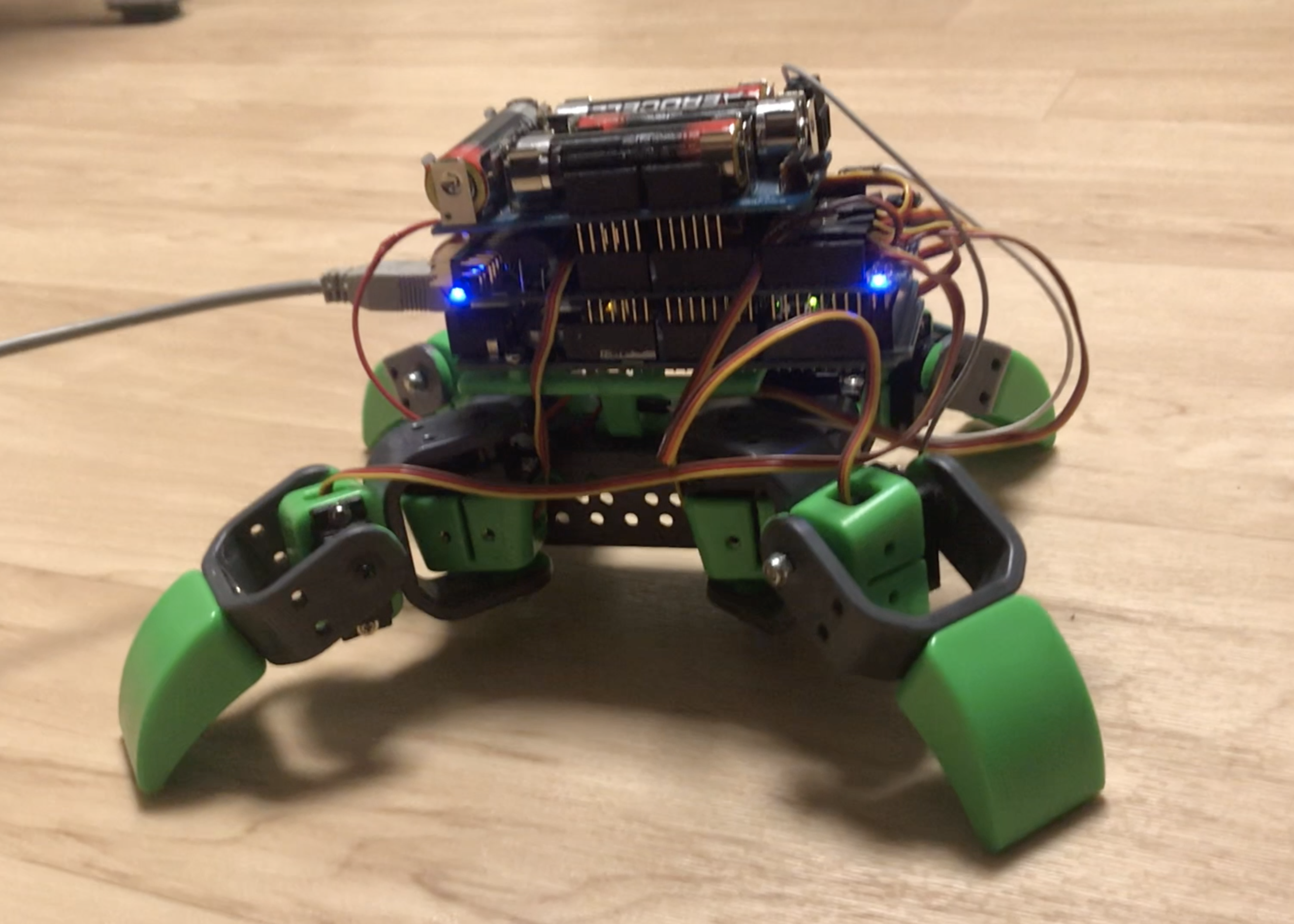}
        \caption{}
    \end{subfigure}%
    \begin{subfigure}{0.45\textwidth}
        \centering
        \includegraphics[width=\textwidth]{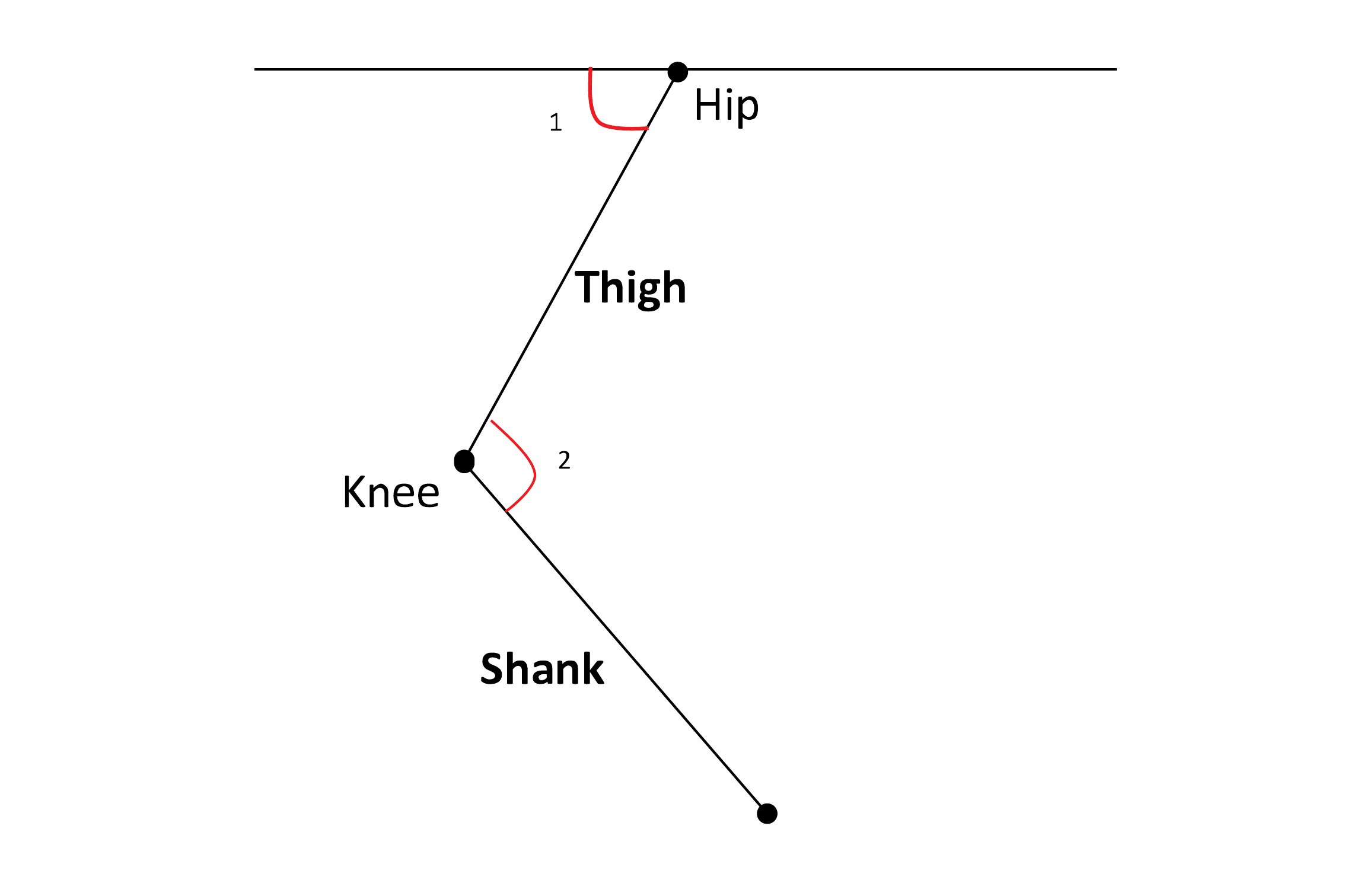}
        \caption{}
    \end{subfigure}
    \caption{(a) ALLBOT VR408. (b) Joints of a single robot leg.}
    \label{fig:allbotmodel}
\end{figure}

We tested our approach on a robot (Allbot VR408 \cite{allbot} shown in figure~\ref{fig:allbotmodel}). The Allbot is a simple quadruped robot with four legs and eight joints that are actuated by standard servo motors. It is equipped with an Arduino Mega board that is used in the experiment to relay motion commands from the network simulation running on the host computer to the servos. The outputs of the motor neurons were used to encode the joint angles. Namely, we used population rate coding, such that the conversion from spikes to angles is done based on:
\begin{equation}
\label{eq:angles}
    \theta_i=\frac{n_{i}}{N_i}*\pi
\end{equation}
where $\theta_i$ is the angle of the joint $i$, $n_{i}$ is the number of spikes of the motor neurons of the joint $i$, and $N_i$ is the total number of neurons representing the $i^{th}$ joint. Each joint is represented by a set of 25 neurons. Activations within a small window of time\footnote{The time window of spike decoding is dependent on the control frequency of the robot} correspond to the angle command. To generate the desired spiking pattern, we used equation~(\ref{eq:angles}) to obtain the number of spikes for each joint $n_{i}$, and generated $n_{i}$ spikes at different randomly chosen neurons encoding joint $i$. The motor activations obtained in this experiment are illustrated in Figure~\ref{fig:robotexp}. Similarly, other control types (such as joint velocity control) can be applied. After learning a predefined motor spiking behavior, the robot is capable of stable walking at different speeds corresponding to different tonic inputs. We provide a video of this experiment\footnote{https://sites.google.com/view/task-independent-cpg/home}. The code for all experiments will be made available upon publication of this paper.

\section{Conclusion}
\label{sec:conclusion}

Spiking CPGs are promising techniques for robot locomotion. Previous methods have either missed the generality feature, scalability, biological plausibility or required parameter hand-tuning. In this work, we were able to present the first general framework for building spiking CPGs that includes all of the desired features. Namely, the framework allows to build CPGs for different types of robots with a minimal design effort, and that is done using a learning algorithm. The used learning method is the remote supervision method which was chosen for its fit to this work's needs and its proved convergence and biological plausibility. Additionally, the resulting CPGs have architectures, behaviors, and features similar to biological counterparts. For instance, they can be modulated by external signals to change the speed of walking even during the process of locomotion. We also showed how our CPG can successfully learn a target spiking behavior allowing it to achieve desired gaits on target robots. A possible extension of this work would be to use reinforcement learning to learn the weights of the network, eliminating the need for behavior demonstration.

\section*{Acknowledgements}
This project/research has received funding from the European Union’s Horizon 2020 Framework Programme for Research and Innovation under the Specific Grant Agreements No. 720270 (Human Brain Project SGA1) and No. 785907 (Human Brain Project SGA2).

\end{document}